\documentclass{article}
\usepackage{fullpage}
\usepackage{hyperref}

\usepackage[utf8]{inputenc} % allow utf-8 input
\usepackage[T1]{fontenc}    % use 8-bit T1 fonts
\usepackage{url}            % simple URL typesetting
\usepackage{booktabs}       % professional-quality tables
\usepackage{multirow}
\usepackage{amsfonts}       % blackboard math symbols
\usepackage{nicefrac}       % compact symbols for 1/2, etc.
\usepackage{microtype}      % microtypography
\usepackage{xcolor}          % colors
\usepackage{listings} % code blocks

\usepackage{dblfloatfix}
\usepackage{mathtools}
\usepackage{graphicx}
\usepackage{siunitx}
\usepackage{multirow}
\usepackage{authblk}

\newcommand{\email}[1]{\href{mailto:#1}{\nolinkurl{#1}}}

\DeclareSIUnit\angstrom{\text {\AA}}

\newcommand{\microsection}[1]{\paragraph{#1}}

\definecolor{ReviewerA}{RGB}{100,80,168}

\definecolor{ReviewerB}{RGB}{21,81,38}

\definecolor{ReviewerC}{RGB}{126,38,64}

\definecolor{ReviewerD}{RGB}{21,114,156}

\definecolor{nequipUpdates}{RGB}{94,16,19}

\definecolor{mayacolor}{RGB}{94,16,19}

\definecolor{matthewcolor}{RGB}{19,16,94}

\definecolor{juliacolor}{RGB}{19,94,16}

\usepackage[numbers,compress,sort]{natbib}
\title{Self-Supervised Learning for Ordered Three-Dimensional Structures}

\author[1]{\small Matthew Spellings}
\author[2]{\small Maya Martirossyan}
\author[2]{\small  Julia Dshemuchadse}

\affil[1]{\footnotesize Work performed at the Vector Institute}
\affil[2]{\footnotesize Cornell University}

\begin{document}

\maketitle

\begin{abstract}

Recent work has proven that training large language models with self-supervised tasks and fine-tuning these models to complete new tasks in a transfer learning setting is a powerful idea, enabling the creation of models with many parameters, even with little labeled data; however, the number of domains that have harnessed these advancements has been limited. In this work, we formulate a set of geometric tasks suitable for the large-scale study of ordered three-dimensional structures, without requiring any human intervention in data labeling. We build deep rotation- and permutation-equivariant neural networks based on geometric algebra and use them to solve these tasks on both idealized and simulated three-dimensional structures. Quantifying order in complex-structured assemblies remains a long-standing challenge in materials physics; these models can elucidate the behavior of real self-assembling systems in a variety of ways, from distilling insights from learned tasks without further modification to solving new tasks with smaller amounts of labeled data \textit{via} transfer learning.

\end{abstract}

\section*{Background}

Recent work on GPT~\cite{brown_language_2020}, BERT~\cite{devlin_bert:_2018}, and related models has proven immensely successful, not only in direct language modeling tasks but also other domains including translation, question answering, and even code~\cite{chen_evaluating_2021} and music~\cite{dhariwal_jukebox_2020} generation. In addition to directly performing transfer learning, prompt engineering has emerged as a promising method to leverage the power of large language models trained on diverse types of texts~\cite{liu_pre-train_2021,sanh_multitask_2022}. The general strategy of pretraining large models on easily-gathered unlabeled data using self-supervised tasks and then fine-tuning on more relevant labeled data is especially appealing for many scientific domains where labeled data may be difficult to come by.

In materials physics, it is well understood how structure plays a significant role in electrical, thermal, or mechanical properties of a material, and scientists target particular structures as they design new materials for desired applications. For crystals, ``structure'' typically refers to the basic building unit which is repeated along a periodic lattice to create a bulk crystal, but---particularly for aperiodic or non-crystalline materials---it can also refer to any symmetry or non-random ordering present in the arrangements of particles or atoms. Assessing order and its evolution in three-dimensional structures is a challenging, but critical method for understanding the self-assembly and growth of complex materials; particularly as the scope and magnitude of experiment and simulation data analysis continues to expand, machine learning techniques that are able to leverage large amounts of unlabeled data will become ever more crucial.

In this work, we use self-supervised learning (SSL) tasks that can broadly be used to train models for quantifying order and distinguishing assemblies in non-idealized material structures. The choice of SSL for this application was inspired by previous work that has developed SSL tasks for three-dimensional point clouds, which are a natural choice for representing three-dimensional positional data. \citet{thabet_self-supervised_2020} formulated self-supervised tasks in terms of a space-filling curve; \citet{sharma_self-supervised_2020} trained deep networks to model data based on a three-dimensional cover tree; the method proposed in \citet{eckart_self-supervised_2021} models simple, soft ``patches'' of 3D point clouds in order to reconstruct its inputs; and \citet{pang_masked_2022} spatially mask patches of point clouds and reconstruct the masked patches using learned networks. However, in contrast to point clouds commonly found in robotics and autonomous vehicles, the types of point clouds discussed here from physics, chemistry, biology, and materials science are comparatively small and based on short-range interactions---often on the order of dozens of points or smaller---and can contain richer information, such as particle type embeddings or atomic features derived from chemical knowledge, rather than solely geometric information. Furthermore, rotation equivariance is required for such physical models for data efficiency and model correctness. In the domain of chemistry, the MolCLR approach~\cite{wang_molecular_2022} uses SSL tasks to learn geometric representations of molecules by pretraining models on molecular graph data using a contrastive loss and graph masking approach. With 3D Infomax~\citep{stark_3d_2022}, networks are optimized to generate 3D information from the latent representation learned from the mutual information between a given molecules  2D molecular graph and its 3D geometric representation. The GEM framework~\citep{fang_geometry-enhanced_2022} uses geometry-aware graph neural networks (GNNs) to predict individual bond distance, angles, and entire distance matrices for a given molecule. The GeoSSL framework~\citep{liu_molecular_2023} focuses on denoising the coordinates of atoms that make up a molecule. These approaches show promise in using self-supervised tasks with geometric data for learning about structural properties of materials---but were designed with chemistry of molecules rather than crystallizing assemblies in mind.

Previous work with more similar applications to ours are self-supervised frameworks operating on periodic materials systems. The Crystal Diffusion Variational Autoencoders~\cite{xie_crystal_2022} method uses SE(3) equivariant GNNs with periodicity to model crystalline unit cells through a variational autoencoding process that can generate stable materials that are optimized for a particular property. The Crystal Twins framework~\cite{magar_crystal_2022} uses GNNs and data augmentations---including masking and adding random noise to coordinates---to formulate a self-supervised learning task. Unlike these examples, the self-supervised tasks presented in this work operate on the basis of individual particles rather than unit cells, and are therefore usable for arbitrary self-assembly data, aperiodic structures such as quasicrystals, or amorphous structures such as glasses and liquid crystals, rather than being restricted to unit cells of crystalline structures. Moreover, while the two frameworks mentioned above specifically target crystalline structures insofar as they are looking for optimizing specific material properties, the relationship between structure and property is well-understood in materials science. The framework we put forth serves an ultimately different function: to study the significantly more mysterious formation and growth of structure and provide insights into how the evolution of order occurs in materials as they form into bulk crystals.

In this work, we present a set of generic tasks applicable to three-dimensional point clouds without any labeling requirements. We train rotation- and permutation-equivariant deep neural networks to perform these tasks using either real self-assembly simulation trajectories or idealized crystal structure unit cells from the AFLOW Encyclopedia of Crystallographic Prototypes \cite{mehl_aflow_2017,hicks_aflow_2019}. We show that we can make use of the embedding-space representations of models trained on ideal structures to analyze self-assembled structures. We evaluate the performance of our self-supervised tasks against existing metrics for quantifying order in materials science, and outperform them in the ability to distinguish structures between highly disordered systems, such as the liquids that form prior to crystallization from two different crystal structures. We believe this is a significant scientific advance in its own right, as there exist virtually no mechanisms in the physical sciences for understanding order in systems of arbitrary liquids or glasses with no symmetry or periodicity. Finally, we demonstrate the breadth of information learned by models from these tasks by performing transfer learning from one self-supervised task to another.

\section*{Methods}

\subsection*{Data sources}

\begin{figure*}[h]
\centering \includegraphics[width=0.6\linewidth]{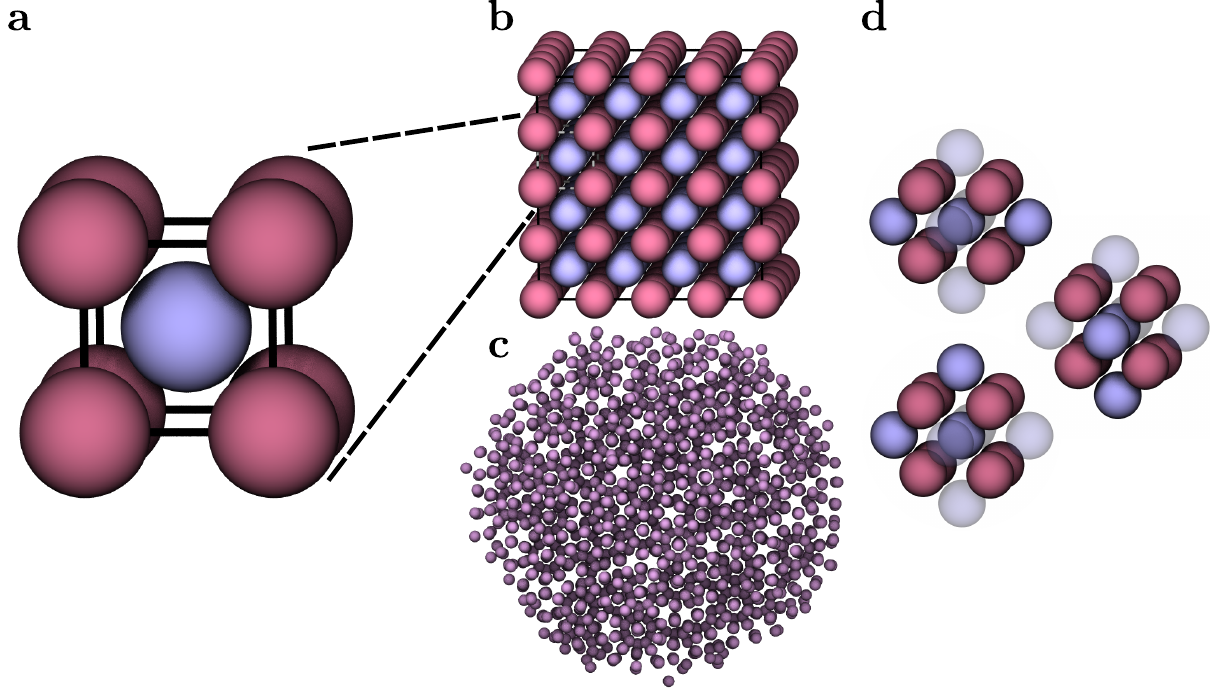}
\caption{
Two types of ordered structural data used in this work.
(a) Periodic three-dimensional systems can be represented by their basic repeat unit, or \textit{unit cell}.
(b) Unit cells can be tiled to fill space; however, non-periodic systems including liquids and (c) quasicrystals have surfaces and may exhibit nontrivial symmetries.
(d) Ordered structures often have different numbers of neighbors at different radial distances. We select the 20 nearest neighbors of each particle to feed into our deep learning models for the tasks presented here. In this diagram, we show three possible permutations of the 10 nearest neighbors around a particle in a CsCl-type crystal structure.
}
\label{fig:crystals}
\end{figure*}

We use two types of data---outlined in Figure~\ref{fig:crystals}---to train and evaluate models on self-supervised tasks in this work. The first type of data used in training are nearly-ideal three-dimensional bulk crystal structures---which are straightforward to generate from the AFLOW~\cite{mehl_aflow_2017,hicks_aflow_2019} database of structures. The second type of data are simulated molecular dynamics assemblies of particles, which are less straightforward to generate without advanced computational resources. Methods used to generate both types of data are explained below.

\microsection{Idealized crystal structure prototypes.} We generate representative samples of bulk crystal structures by selecting one-component unit cells of 3D structures from the AFLOW Encyclopedia of Crystallographic Prototypes~\cite{mehl_aflow_2017,hicks_aflow_2019} (listed in Appendix~\ref{app:structure_prototype_list}) and adding random noise to the coordinates after replicating the unit cells along a crystal lattice. An example unit cell and replicated periodic crystal structure are shown in Figure~\ref{fig:crystals}(a) and (b), respectively. Because crystal structures typically consist of successive shells of highly symmetric arrangements of particles with the same radial distance, adding random noise to the coordinates allows us to sample the permutations of particles from the farthest neighbor shells of a given structure for a fixed number of nearest neighbors. We show a few such sample permutations in Figure~\ref{fig:crystals}(d).

\microsection{Self-assembly data.} Unit cells are easy to generate and because of this, can provide valuable information about the bulk behavior of materials. Unfortunately, many types of simulations do not readily yield such idealized data: particles can form disordered or poorly-ordered structures, they can contain grain boundaries and other defects, or expose distinct surface structures. Moreover, aperiodic structures cannot be generated with a unit cell, as they do not have periodicity in three dimensions. An icosahedral quasicrystal---which has rotational symmetry but is aperiodic---is shown in Figure~\ref{fig:crystals}(c). In addition, self-assembly data allows for the study of \textit{how} ordered structures are formed from the fluid phase, which idealized structures cannot provide. To account for these more realistic configurations of particles, we perform molecular dynamics simulations of systems that are thermalized and slowly cooled using HOOMD-blue~\cite{howard_efficient_2016,anderson_hoomd-blue_2020}. Details of particle interactions and simulation dynamics are described in Appendix~\ref{app:md_simulation_params} and full trajectories are available in the Supplementary Information.

\subsection*{Geometric algebra attention networks}

One of the key architectural requirements of performing transfer learning on the variety of geometric, semantic, rotation-equivariant, and rotation-invariant tasks we present here is control over the equivariance of the signals passing through the network; to that end, we build flexible deep learning architectures by composing geometric algebra attention layers~\cite{spellings_geometric_2021}. These layers attain rotation equivariance by deriving their geometric calculations from the geometric algebra, as well as permutation equivariance through an attention mechanism. For this work, we utilize pairwise attention; briefly, this consists of computing the geometric products $p_{ij}$ between all pairs of input multivectors\footnote{In the geometric algebra in three dimensions, multivectors are a linear combination of scalars, vectors, bivectors, and trivectors---representing 0, 1, 2, and 3-dimensional quantities, respectively---so that vectors are a subset of multivectors; here, we use networks with multivector intermediates and convert input vectors to their multivector numerical representation as the first step in the network.} $\vec{r}_i$ (with associated type embeddings $v_i$\footnote{In this work, we only analyze single-component systems, but we retain the type embeddings as network inputs to account for additional particle features and multicomponent systems in the future. Within the network, the values $v_i$---the rotation-invariant representations of the geometry of each neighboring bond $i$---are learned.}) in a given point cloud. The pairwise geometric products are converted into a geometric embedding \textit{via} their rotation-invariant attributes (which include magnitudes of vectors and bivectors, as well as values of scalars and trivectors---represented generally for any multivector by the \textit{invariants} function) using a multilayer perceptron (MLP) $\mathcal{V}$. In this work we generate a full set of rotation-invariant attributes including all input vectors forming the product, rather than only using the rotation-invariant attributes of the final product as in reference~\cite{spellings_geometric_2021}. The geometric embedding is combined with a node-level signal summary of the pair of input points (the rotation-invariant attributes passed in as inputs to the network are simply particle type embeddings for both particles in the bond); the summary representation is derived from a learned linear projection $\mathcal{M}(v_i, v_j) = A \cdot \text{concatenate}(v_i, v_j)$ using a learned linear projection $\mathcal{J}(v_{geometry}, v_{nodes}) = B \cdot \text{concatenate}(v_{geometry}, v_{nodes})$, which are passed through another MLP $\mathcal{S}$ to generate attention score logits. A new rotation-invariant output $v_i^\prime$ or rotation-equivariant output $\vec{r}_i\!^\prime$ (using a rescaling MLP $\mathcal{R}$ and learned scalar constants $\alpha_n$) can be calculated as follows:

\begin{align}
\label{eq:basic_attention}
\begin{split}
 p_{ij} &= \vec{r}_i\vec{r}_j \\
 q_{ij} &= \text{invariants}(\vec{r}_i, \vec{r}_j , p_{ij}) = (|\vec{r}_i|, |\vec{r}_j|, \vec{r}_i\cdot \vec{r}_j, |\vec{r}_i \wedge \vec{r}_j|) \\
 v_{ij} &= \mathcal{J}(\mathcal{V}(q_{ij}), \mathcal{M}(v_i, v_j)) \\
 w_{ij} &= \operatorname*{\text{softmax}}\limits_{j}(\mathcal{S}(v_{ij})) \\
 v_i^\prime &= \sum\limits_{j} w_{ij} v_{ij} \\
\vec{r}_i\!^\prime &= \sum\limits_{j} w_{ij} \mathcal{R}(v_{ij})\left(\alpha_0 \vec{r}_i + \alpha_1 \vec{r}_j + \alpha_2 \vec{r}_i\vec{r}_j \right) .
\end{split}
\end{align}

We note that the formulas in Equation~\ref{eq:basic_attention} produce a permutation-equivariant result: for each input bond $(\vec{r}_i, v_i)$, a value ($\vec{r}_i\!^\prime$ or $v_i^\prime$) is produced as an output. This is suitable for composing multiple layers in a deep learning context. To instead produce a permutation-invariant result---which creates a single summary value for an entire point cloud of bonds---the softmax and summation are applied over $i$ and $j$, rather than only over $j$. For the work here, we use point clouds of fixed size---obtained by finding the 20 nearest neighbors in space around each particle---although the attention-based framework is flexible and supports heterogeneous numbers of neighbors for each particle, as would be expected when calculating neighbors within fixed cutoff distance or using a Voronoi tessellation. Further details of normalization layers and other hyperparameters---including code to build and train models on the datasets used here---are available in Appendix~\ref{app:hyperparameters} and the Supplementary Information.

\subsection*{Self-supervised tasks}

\begin{figure*}[h]
\centering \includegraphics[width=0.75\linewidth]{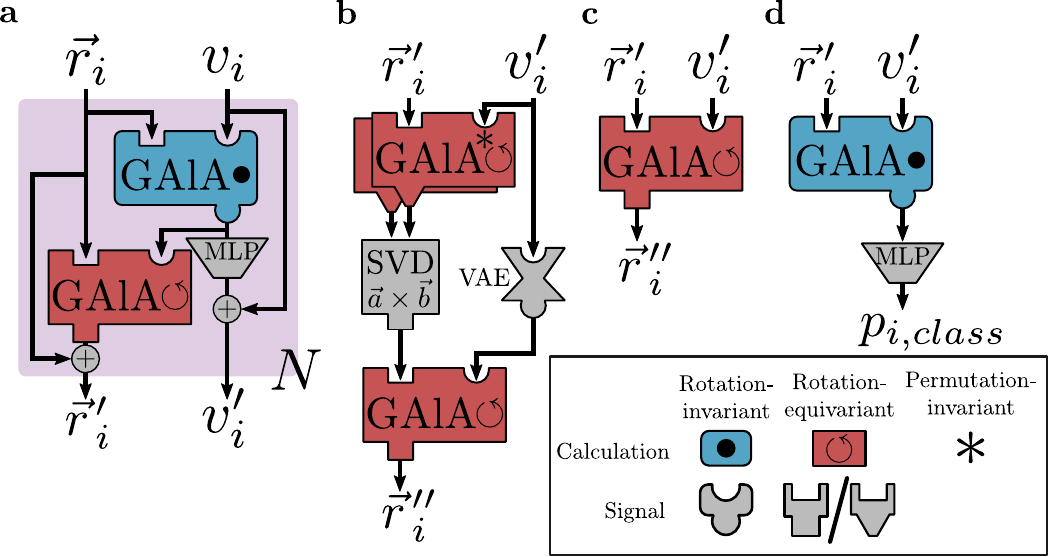}
\caption{Architectures for equivariant, self-supervised transfer learning tasks.
(a) The core shared by all architectures consists of a series of learned, rotation- and permutation-equivariant Geometric Algebra Attention (GAlA) blocks propagating one rotation-invariant and one rotation-equivariant signal for each input point. Architectures used in this work stack $N=3$ sets of layers inside the core.
(b) The head layers for the autoencoding task produce two vectors using geometric algebra attention, which is converted into a bottleneck value specifying an orientation of the point cloud \textit{via} SVD and a cross product. The rotation-invariant values are passed through a standard variational autoencoder~\cite{kingma_auto-encoding_2014}. Finally, geometric algebra attention between the per-input embeddings and the orientation vectors generates the output value.
(c) The architecture for denoising simply adds a layer of rotation-equivariant, permutation-equivariant attention. For shift identification and nearest bond regression, we instead use a permutation-invariant final attention layer.
(d) For classification tasks (frame classification and noisy bonds), a simple MLP is applied to generate logits for classification. Frame classification uses a final permutation-invariant reduction over the point cloud before the MLP, while the noisy bond architecture uses permutation-equivariant attention.
}
\label{fig:architectures}
\end{figure*}

As detailed in Figure~\ref{fig:architectures}, we use a common core architecture for all self-supervised tasks. This architecture takes in a point cloud of nearest-neighbor bonds and associated type embeddings and refines two types of signals in parallel: a rotation-invariant value associated with each bond, alongside a rotation-equivariant multivector for each bond. Using full multivector intermediate geometric values within the network and use of learned rotation-invariant and -equivariant transformations are two improvements over the architectures shown in reference~\cite{spellings_geometric_2021}. Additional task-specific layers are added to the core architecture to specialize it for each task; for example, a simple MLP can be added to the rotation-invariant signal for each bond to classify bonds, or the vector component can be extracted from the rotation-equivariant output to produce a denoising autoencoder. For geometric tasks producing points or point clouds as output, we train networks using the mean square error loss; for classification tasks, we use categorical crossentropy.

\microsection{Autoencoding.} Autoencoders have a rich history of usage in machine learning for many years and can be seen as the least constrained self-supervised task because they impose no restrictions on the structure of their input data. However, this freedom of problem specification is counterbalanced by complexity in model formulation: autoencoder models must usually be designed to restrict information in some way to prevent models from learning trivial identity functions. Using autoencoding-based strategies to initialize the weights of deep neural networks was a key area of research in the development of modern deep learning~\cite{erhan_why_2010,vincent_stacked_2010}. Pretraining \textit{via} greedy autoencoding-based strategies bears many similarities to the approaches described in this work, however the autoencoders presented in this work are trained end-to-end, rather than locally for each layer, before potentially performing transfer learning on another task. The work presented here is also focused on a broader set of tasks than solely autoencoding. Variational autoencoders~\cite{kingma_auto-encoding_2014} are a popular architecture imposing additional constraints on the embedding space learned by the model. In this work, we train variational autoencoders to reproduce input point clouds. As a bottleneck, we produce two vectors, calculate their singular value decompositions, and take the cross product to produce an orthonormal set of basis vectors. We also produce a rotation-invariant signal that is passed through a variational autoencoder to generate a smooth embedding space. Using the orientation information from the orthonormal basis set and the geometric information from the embedding space, we can train a decoder to reconstruct the input point cloud using a mean square error loss on each coordinate dimension.

\microsection{Denoising.} Denoising autoencoders are one of the simplest adaptations of autoencoders to lessen the difficulty of models learning too-simplistic identity functions. We teach models to produce an original point cloud from a perturbed version of the same point cloud using the mean square error loss. The perturbation applied to each coordinate of each point is generated from a Gaussian distribution with standard deviation $0.5 \sigma_0$. A rigid rotation and translation are applied to the perturbations of the entire point cloud to remove any net rotation or translation from the random noise. The difficulty of this task can be tuned by adjusting the magnitude of the added noise. We use a single permutation-equivariant, rotation-equivariant attention layer atop the core as the architecture for the denoising task.

\microsection{Frame classification.} In many cases when studying physical systems, we have additional information that could carry meaningful signal when used as labels. Individual frames from a self-assembly trajectory could be fed in with observables/physical conditions as a label to allow models to distinguish structures as a function of a particular observable/physical condition. For self-assembly simulations, we use the physical condition of temperature, synonymous with time elapsed in the cooling simulation for the frame classification task. For the crystal structure prototypes drawn from the database, we use the structure type as a label, as shown in Figure \ref{fig:embeddings}. We use a typical classification MLP on top of a permutation-invariant attention reduction over all of the bonds in the point cloud to generate per-class logits.

\microsection{Shift identification.} For this task, we shift all bonds of an input point cloud by a Gaussian-distributed random vector with a standard deviation of $0.5\sigma_0$ and train models to predict this displacement vector. Although this task would be trivial for most crystal structures given clean neighbor shells due to their high symmetry, for a large set of structures with differing neighbor shell sizes, we find this task to extract useful information about structure. The difficulty of this task can be adjusted by tuning the magnitude of the displacement: larger-magnitude displacements are easier to predict, while smaller displacements can be washed out by the disorder inherent in the structures under consideration. As shown in Figure~\ref{fig:architectures}(c), we use a permutation-invariant, rotation-equivariant attention reduction over the bonds in the point cloud to predict the displacement vector.

\microsection{Noisy bond classification.} This task randomly adds noise (Gaussian distributed, with standard deviation $0.5\sigma_0$) to half of the bonds of each point cloud. A permutation-equivariant classifier is then trained to identify, for each bond, whether it was perturbed or not. The difficulty of this task depends on how disordered or complex the structures under consideration are, and the magnitude of the noise that is added to bonds. As shown in Figure~\ref{fig:architectures}(d), we use a permutation-equivariant attention calculation, followed by an MLP producing two class logits, to classify bonds as perturbed or pristine.

\microsection{Nearest bond regression.} This task is most similar to the masked language modeling~\cite{devlin_bert:_2018,brown_language_2020}, point cloud~\cite{pang_masked_2022}, and graph-level~\cite{wang_molecular_2022} self-supervised tasks: for each point cloud, we remove the nearest bond from the set of inputs, and teach the model to predict that bond. We choose the nearest bond, to decrease degeneracy of the problem space; a randomly-selected bond could choose particles that are not part of a full neighbor shell, which would require a multimodal output for every remaining missing particle in the neighbor shell, as shown in Figure~\ref{fig:crystals}(d). As shown in Figure~\ref{fig:architectures}(c), we use a permutation-invariant, rotation-equivariant attention reduction over the bonds in the point cloud to predict the nearest bond for the point cloud.

\section*{Results}

\subsection*{Mapping structure space through learned embeddings}

\begin{figure*}[h]
\centering \includegraphics[width=0.99\linewidth]{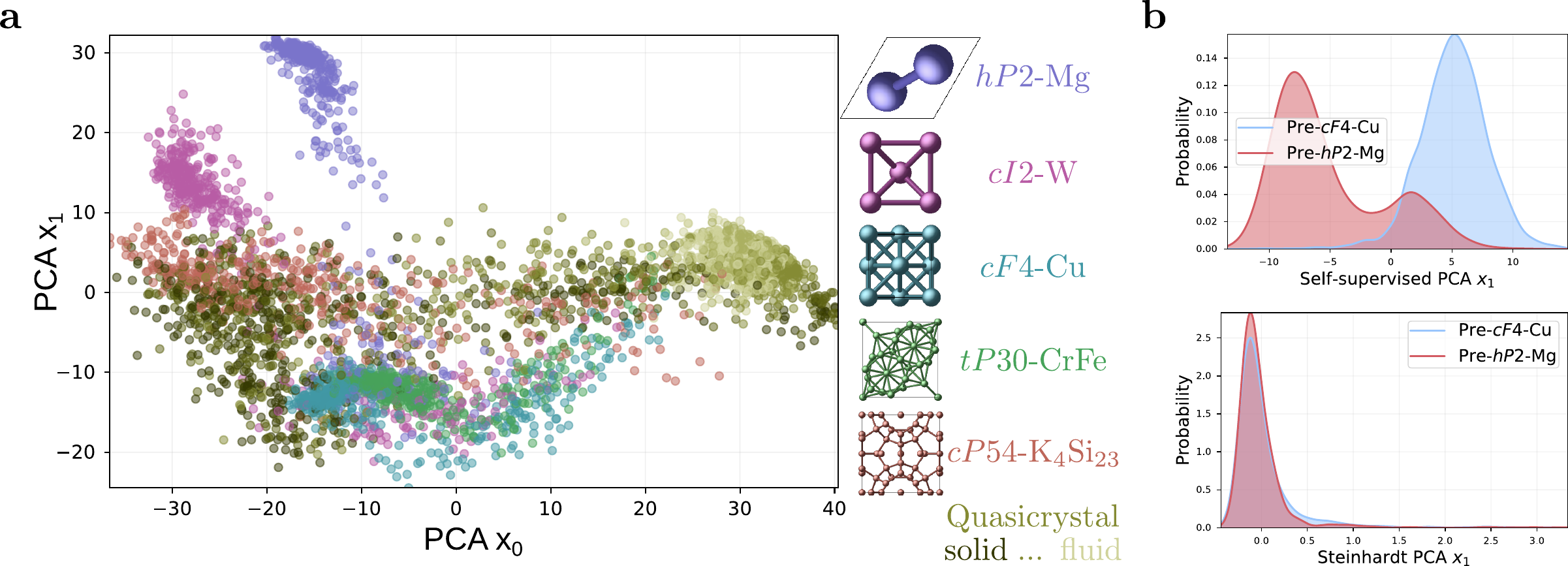}
\caption{
(a) Two-dimensional projection of embeddings for a set of self-assembled structures using a self-supervised model trained on bulk unit cell data. The generated embeddings hold useful information about three-dimensional structure, as they show distinct signatures for the different structures and have a reasonable distribution even for crystalline surfaces and the icosahedral quasicrystal, which were not included in the training set: a capability which is necessary for models to generalize well for out-of-distribution observations.
(b) Representations learned by self-supervised models can perform better than classical methods: here, embeddings learned for a frame classification model on solids can distinguish between pre-crystallization liquids significantly better than the Steinhardt $Q$ order parameters are capable of.
}
\label{fig:embeddings}
\end{figure*}

We can use embeddings to understand how well the models trained by self-supervised tasks are able to distinguish between different types of crystal structures. We first train a network on the frame classification task with crystal structure prototype data, then evaluate the learned internal representation of the model on self-assembly simulations. The Principal Component Analysis (PCA)~\cite{pearson_lines_1901} projection of the data in two dimensions is plotted in Figure \ref{fig:embeddings}(a), showing that the model is able to learn reasonable representations of different self-assembled structures even though it was trained on idealized unit cell data. In addition, the embedding for a cooling trajectory of an icosahedral quasicrystal is included, which is not in the training set given that there is no unit cell that can generate this aperiodic structure. Although the quasicrystal---like the periodic crystal structures---consists of locally symmetric motifs, they do not tile space periodically. We use the quasicrystal as a sample structure that is significantly different from the structures used to train the self-supervised model, to show that reasonable embeddings are generated for structure types the model has not seen before.

To quantitatively compare the embeddings generated through self-supervised tasks, we select a few different simulation snapshots showcasing phase transitions and different structures and use the generated embeddings as \textit{order parameters} to distinguish between each pair of structures in a binary classification task. For each self-supervised task, as well as three other methods that are standard for materials systems---described in Appendix~\ref{app:featurization_baselines}---we evaluate featurizations in their capacity to distinguish gas ($G$), liquid ($L$), and solid ($S$) structures for self-assembly systems that will form the same final structure ($A/A$) or two different structures ($A/B$) at low temperature. We evaluate the featurizations and use the first principal component as a score to distinguish between the two snapshots being evaluated, and quantify the classification capability of each representation through the area under the curve of a receiver operating characteristic~\cite{hanley_method_1983} in a form of zero-shot learning in Table~\ref{tab:zero_shot_aucs}. While all methods behave sensibly---distinguishing two identical liquids ($L_A/L_A$) should perform poorly, and distinguishing a gas from a liquid ($G_A/L_A$) should perform well---the other phase transitions showcase the relative strengths and weaknesses of the representations. Identifying liquid-solid transitions is one of the main purposes for the Steinhardt $Q$~\cite{steinhardt_bond-orientational_1983} and neighbor-averaged spherical harmonic~\cite{spellings_machine_2018} representations, which have the advantage of strong symmetry priors enforced by use of spherical harmonics in their featurizations, so it is no surprise that these methods most accurately identify the crystallization process. The arguably most difficult task in Table~\ref{tab:zero_shot_aucs} is distinguishing between liquids that will form two different structures, $L_A/L_B$. In this case, the learned representations---with the exception of the noisy bond identification task---outperform all existing methods. In the scientific literature, this task has virtually no precedent; while \citet{hu_revealing_2022} use the Steinhardt order parameter to distinguish between competing liquid motifs prior to crystallization, they tune parameters carefully for the specific liquid structures under investigation and it is not obvious that this method can be generalized easily. The frame identification task, shown as a kernel density estimate in Figure~\ref{fig:embeddings}(b) alongside the Steinhardt embedding density, performs the best here---despite being trained \emph{only} on perturbed ideal crystalline unit cells---and the geometric algebra attention networks are able to to selectively transmit more useful information about each bond than the averaged representations over all bonds as the existing spherical harmonic-based methods do.

\begin{table}[h]
\caption{ROC curve AUC comparison of representations for self-assembly systems. The six SSL tasks presented here (Frame classification, noisy bond classification, autoencoding, denoising autoencoding, nearest bond regression, and shift identification) as well as three standard materials methods (Steinhardt, local neighbor-averaged spherical harmonics, and radial distance) are used to distinguish pairs of snapshots. Gases ($G$), liquids ($L$), and solids ($S$) that will form two different structures ($A/B$) or the same structure ($A/A$) are compared. Detailed information about the simulations and methods used can be found in Appendix \ref{app:featurization_baselines}.}
\label{tab:zero_shot_aucs}
\centering \tabcolsep=0.11cm \begin{tabular}{cccccccccc}\toprule
 & Frame & Noisy & Autoencoder & Denoising & Nearest & Shift & $Q_\ell$ & $|\Psi|$ & $\frac{|r_n|}{|r_0|}$
\\ \midrule
$L_A$/$L_A$ & 0.512 & 0.509 & 0.512 & 0.505 & \textbf{ 0.513 } & 0.509 & 0.504 & 0.504 & 0.509
\\ \midrule
$G_A$/$L_A$ & \textbf{ 1.000 } & \textbf{ 1.000 } & \textbf{ 1.000 } & 0.918 & 0.999 & \textbf{ 1.000 } & 0.999 & 0.998 & 0.972
\\ \midrule
$L_A$/$S_A$ & 0.726 & 0.815 & 0.570 & 0.579 & 0.537 & 0.584 & 0.903 & \textbf{ 0.955 } & 0.752
\\ \midrule
$G_A$/$S_A$ & 0.957 & 0.958 & 0.955 & 0.805 & \textbf{ 0.969 } & 0.965 & 0.949 & 0.897 & 0.909
\\ \midrule
$L_A$/$L_B$ & \textbf{ 0.962 } & 0.530 & 0.859 & 0.814 & 0.690 & 0.622 & 0.506 & 0.505 & 0.530
\\ \midrule
$S_A$/$S_B$ & \textbf{ 1.000 } & 0.845 & 0.895 & 0.828 & 0.644 & 0.704 & 0.750 & 0.788 & 0.582
\\ \bottomrule
\end{tabular} \end{table}

Embeddings such as these can be used to qualitatively understand the structural behavior of systems through visual analysis. They could easily be made quantitatively useful through 
a variety of means depending on the application, including diffusion maps~\cite{coifman_diffusion_2006} for studying the manifold formed by the data, probability density modeling through normalizing flows~\cite{rezende_variational_2015} or other methods, or as the basis for transfer learning---which we showcase in the next section.

\subsection*{Transfer learning between tasks}

Inspired by recent work using networks initially trained on language modeling tasks to perform a variety of other language-related functions, we probe how well models trained on one structural task perform on other tasks with a limited amount of new labeled training data. As a more systematic, quantifiable stand-in for arbitrary new labeled data---which could come from physical experiments, expensive simulations, or other means---we perform transfer learning between all pairs of tasks listed in the Methods section using crystal structure prototype data. Examples of relevant downstream tasks include analyzing structures of liquid-liquid separation in cells~\cite{chong_liquidliquid_2016}, proteins~\cite{andre_liquidliquid_2020}, or kinetic pathways to crystallization via competing liquids~\cite{zhang_nonclassical_2017, hu_revealing_2022}. We first train networks on a \textit{source} task, then reuse the core of those trained networks as an initial state to train networks on a \textit{target} task. For retraining the networks on the target task, we limit the training data to a new randomly-chosen subset of the available data. We show results for both allowing the core weights to be updated, as well as ``freezing'' the core weights and only retraining the non-core portions of the network. For each pair of tasks, we show the best validation set performance (the mean absolute error for geometric tasks, or $1 -\text{accuracy}$ to keep lower values better for classification tasks) achieved on the target task as a function of the fraction of available training data for fine-tuning.

\begin{figure*}[h]
\centering \includegraphics[width=0.99\linewidth]{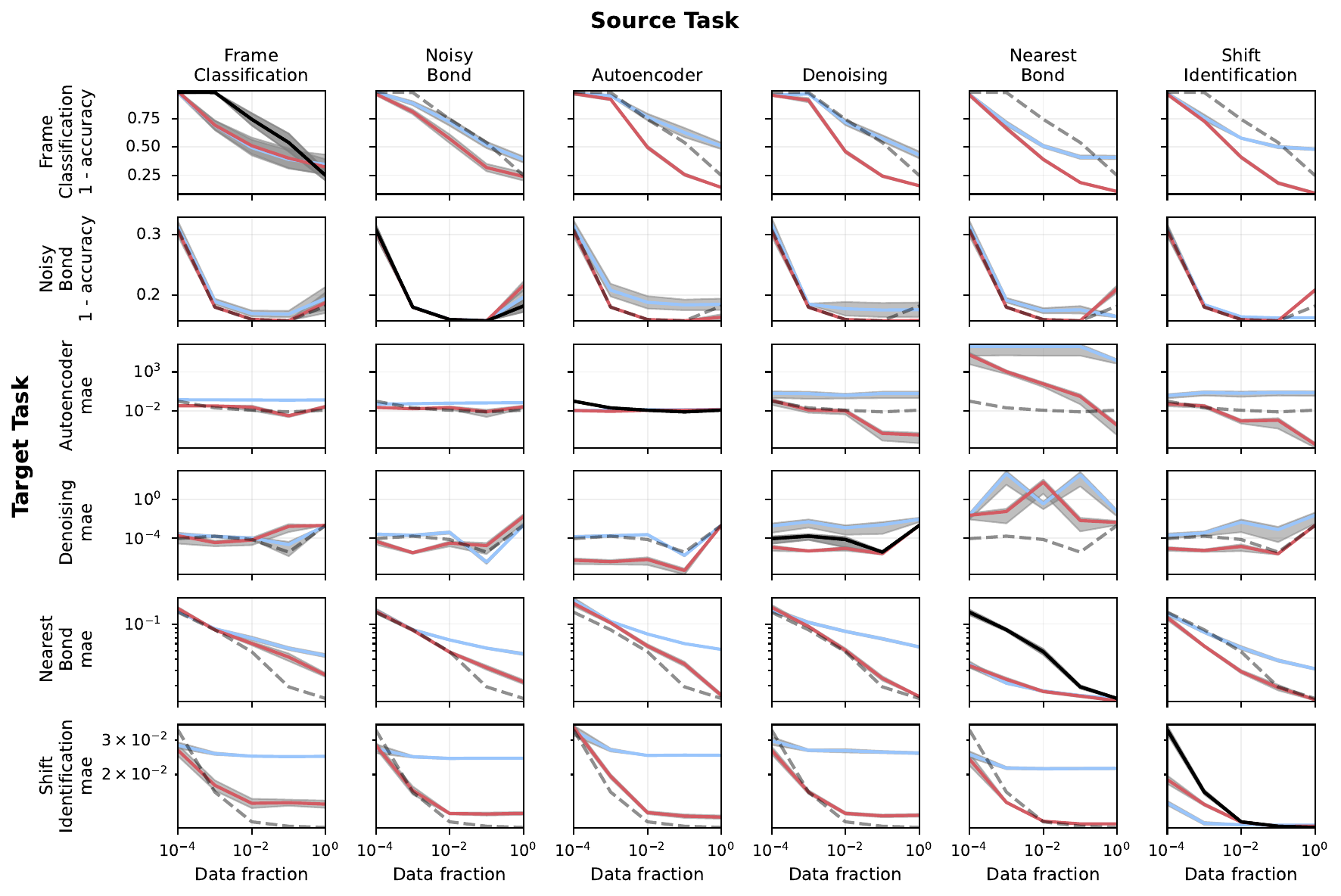}
\caption{
Transfer learning performance of rotation-equivariant deep neural networks. Models are first trained on all  idealized crystal structure unit cell data available for a \textit{source} task, then fine-tuned on a limited amount of data for a \textit{target} task. Blue lines indicate performance for cases where the ``core'' weights of the network are not updated during retraining; red lines indicate performance for updating all weights, and black lines indicate direct performance of pure (non-transfer learning) models with limited training data. For ease of comparison to transfer learning results, results from direct training on limited data are reproduced in off-diagonal plots as dashed lines. Gray-shaded areas indicate one standard error of the mean over 10 independent training replicas. Data are available in numerical form in Appendix~\ref{app:xfer_tables}.
}
\label{fig:transfer_learning}
\end{figure*}

As shown in Figure~\ref{fig:transfer_learning} and Appendix~\ref{app:xfer_tables}, in many cases, models trained on different tasks learn to extract different types of information from their training data, leading to improved performance compared to directly training for the given target task (indicated by black dashed lines). In particular, most pretraining tasks are helpful for frame classification: by adding a single attention layer and MLP to the pretrained core, most networks attain higher accuracy than networks trained solely on the frame classification task. This makes sense because frame classification of different structures is the most abstract, least geometrically precise task among those described here; the higher-dimensional labels for most of the self-supervised tasks allow the networks to extract more information that is then applicable to the less-precise task of frame classification. In contrast, the nearest bond and shift identification tasks seem to learn fairly distinct localized, lower-level representations of the data: in most cases, learning on other tasks is not substantially helpful for these cases, aside from retraining a shift identification network on the nearest bond task. The broad applicability and mutual similarity of the autoencoder and denoising autoencoder tasks shows through their competitive performance on other tasks and usefulness in pretraining against each other, respectively; however, some networks---especially those pretrained on the highly localized nearest bond task---seem to have difficulty converging to a stable, low-error state when retraining on the autoencoder task with smaller datasets. In contrast, networks pretrained on the nearest bond and shift identification tasks and fine-tuned on the autoencoder task with the full training dataset perform surprisingly well, demonstrating that domain-specific complementary, but related, self-supervised objectives can be helpful even for very generic tasks given an abundance of data. We would expect that utilizing all self-supervised representations simultaneously---for example by creating a new ``core'' architecture that concatenates the learned representation from each self-supervised task and applies the final layers as usual---would perform well on all tasks; however, due to constraints of hardware capacity, we leave this architecture for future work. Overall, by comparing the performance of networks trained with the core weights frozen (blue lines) and with pretrained weights solely used as initialization (red lines), we can see that, in most cases, the tasks are sufficiently distinct that fundamental changes to the earlier representations learned in the network are helpful when adapting to a new task. This agrees with findings in large networks trained to model languages, where earlier layers learn lower-level representations of their inputs~\cite{rogers_primer_2021}.

\section*{Conclusion}

While generating large amounts of data for self-assembly simulations is more computationally feasible now---the study in \citet{dshemuchadse_moving_2021} consisted of over 8,500 simulations with thousands of particles in each simulation---understanding the fine-scale structural behavior of these systems is a much more complex task. Historically, automated analysis of three-dimensional structural data has been restricted to fairly low-dimensional representations~\cite{steinhardt_bond-orientational_1983,boattini_unsupervised_2019} or shallow machine learning on hand-crafted featurizations~\cite{spellings_machine_2018}. Self-supervised methods such as those presented here harness the power of geometric deep learning to build powerful, generic feature learners that can yield insight into the behavior of systems on a much finer scale than would be feasible through manual analysis. By designing deep learning architectures for physically-relevant data symmetries, we can begin to apply standard deep learning tools such as analysis of internal representations to solve new types of problems. Although we have focused primarily on the rotation-invariant signal passing through a single point in the networks thus far, future work could quantify the types of information passing through the geometric (rotation-equivariant) channels within the network, as well as learned attention weights throughout the network in order to identify which particles and signals are most important for model predictions.

In summary, we have presented rotation- and permutation-equivariant neural network architectures and a collection of self-supervised tasks suitable for analyzing ordered and disordered three-dimensional structures. Particle-centered point cloud representations of materials are useful in understanding many types of non-ideal structures that can not be represented using unit cells, such as disordered and aperiodic systems, and we hope that the datasets and tasks presented here motivate new architectural work on these data structures. The novel architectures and models presented here can provide new insights to self-assembly and structure formation by applying self-supervised learning tasks, analyzing the embedding-space representations, or using transfer learning to improve performance, all of which can help leverage deep learning to solve new types of problems in materials science.

% don't uncomment during review
%\section*{Acknowledgments}
%
%Resources used in preparing this research were provided, in part, by the Province of Ontario, the Government of Canada through CIFAR, and companies sponsoring the Vector Institute (\url{www.vectorinstitute.ai/partners}). M.\ M.\ M.\ acknowledges support from the National Science Foundation Graduate Research Fellowship Grant No. DGE-1650441 (2019-2021) and DGE-2139899 (2021-2022) and from the Dolores Zohrab Liebmann Fund Fellowship. The authors declare no competing financial interests.

\bibliography{references}

\clearpage

\appendix

\section{Frame classification for phase identification}

We show direct application of the frame classification task to analyze self-assembling systems and identify phase behavior of a cooling system. Although capturing the gas--liquid phase transition is straightforward through the use of the local density, capturing transitions from liquid to crystal for arbitrary structures can be much more difficult; while physicists have used methods such as the Steinhardt bond-orientational order parameters~\cite{steinhardt_bond-orientational_1983}, these typically require a fairly detailed understanding of the structure that will be formed in order to robustly analyze systems~\cite{mickel_shortcomings_2013}. Instead, we can use this SSL task to automatically find the distinct phases in a simulation without needing to interpret or understand the structure that is being assembled.  We do this by taking every fourth frame from a trajectory as training data and training a classifier to associate the structures found in each frame---each corresponding to a distinct temperature---to that particular frame. We analyze the phase behavior by visualizing the histogram of classification predictions found in frames excluded from training---in this particular case, we evaluate predictions for each frame immediately after every training set frame.

\begin{figure*}[h]
\centering \includegraphics[width=0.95\linewidth]{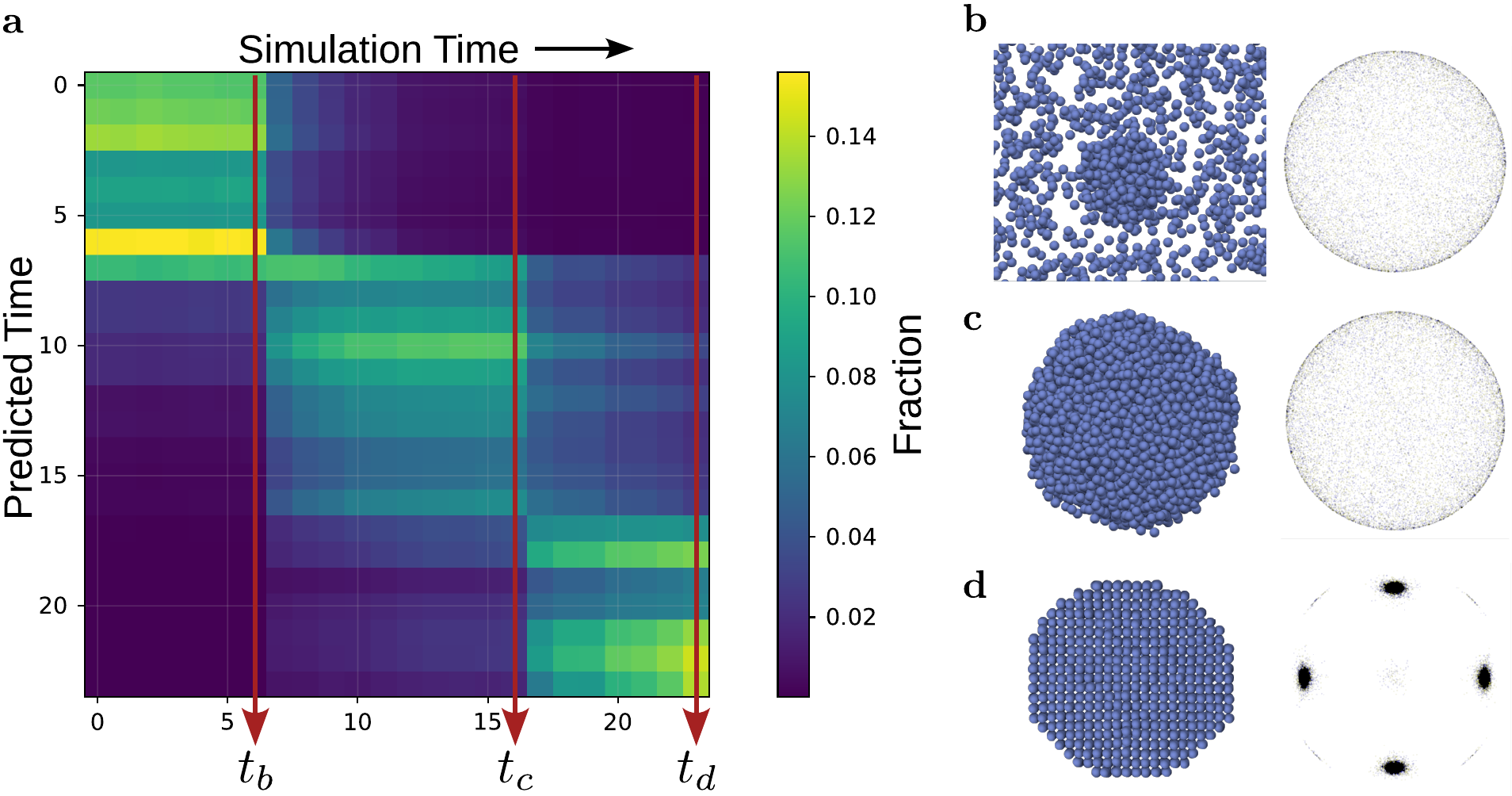}
\caption{(a) Histogram of predicted frames over a simulated cooling trajectory of particles forming the $cI2$-W (``body-centered cubic'') crystal structure. Models trained on simple self-supervised tasks can help automate the application of human knowledge, such as identifying distinct gas, liquid, and solid phases here.
(b--d) System snapshots and bond-orientational order diagrams~\cite{dzugutov_formation_1993} (BOODs) showing the development of order for the gas (b), liquid (c), and solid (d) phases at the times indicated by red arrows. Peaks in BOODs indicate global alignment of bonds and can serve as a helpful fingerprint to identify crystallization.
}
\label{fig:phase_classification}
\end{figure*}

As shown in Figure~\ref{fig:phase_classification}, even this simple self-supervised analysis can quickly help us understand the dynamical phase behavior of self-assembling systems. Incorporating our understanding of the phase behavior of materials as a function of temperature, we can identify gas--liquid, gas--solid, and liquid--solid phase transitions at a glance for large numbers of trajectories.

\section{Crystallographic prototypes used for training}
\label{app:structure_prototype_list}

We use multiple replicas of unit cells---each replicated to at least 4096 particles and with different levels of Gaussian noise (with standard deviations of $10^{-2}\sigma_0$, $3\cdot 10^{-2}\sigma_0$, and $5\cdot 10^{-2}\sigma_0$) applied to each coordinate---to emulate thermal noise typically observed in real self-assembled systems. Thermal noise is important not only to provide a more realistic distribution of bond distances and angles for the networks to learn, but also to sample the different permutations of ways to choose neighbors. Fifty different single-component crystal structure prototypes are taken from the AFLOW Encyclopedia of Crystallographic Prototypes~\cite{mehl_aflow_2017,hicks_aflow_2019} for use in self-supervised training. Structures are named based on their crystal system (indicated by the first character) and unit cell centering (indicated by the second character)---which together denote the Bravais lattice, the number of particles in the unit cell, and a representative real-world material that forms the given structure.

\begin{table}[h]
\caption{List of crystal structure prototypes used for pretraining in this work.}
\label{tab:pyriodic_structures}
\centering \begin{tabular}{ccccc} \toprule
$aP4$-Cf & $cF136$-Si & $cF4$-Cu & $cF8$-C & $cI16$-Li \\ \midrule
$cI16$-Si & $cI2$-W & $cI58$-Mn & $cP1$-Po & $cP20$-Mn \\ \midrule
$cP46$-Si & $hP1$-HgSn$_{6-10}$ & $hP2$-Mg & $hP3$-Se & $hP4$-La \\ \midrule
$hP6$-C & $hP6$-Sc & $hR1$-Hg & $hR1$-Po & $hR105$-B \\ \midrule
$hR12$-B & $hR2$-As & $hR2$-C & $hR3$-Sm & $mC12$-Po \\ \midrule
$mC16$-C & $mC34$-Pu & $mP16$-Pu & $mP32$-Se & $mP4$-Te \\ \midrule
$mP8$-C & $mP84$-P & $oC4$-U & $oC8$-Ga & $oC8$-P \\ \midrule
$oF128$-S & $oF8$-Pu & $oP16$-C & $oP8$-Np & $tI16$-S \\ \midrule
$tI2$-In & $tI2$-Pa & $tI4$-Si & $tI4$-Sn & $tI8$-C \\ \midrule
$tP12$-C & $tP12$-Si & $tP30$-U & $tP4$-Np & $tP50$-B \\ \bottomrule
\end{tabular}
\end{table}

\section{Molecular dynamics simulation details}
\label{app:md_simulation_params}

 Particles interact \textit{via} isotropic Oscillatory Pair Potential (OPP) and Lennard-Jones--Gauss (LJG) interactions---described in reference~\cite{dshemuchadse_moving_2021}---with potential energy functions:

\begin{align}
\label{eq:opp_ljg_potentials}
\begin{split}
U_\textrm{OPP} &= \frac{1}{r^{15}} + \frac{\text{cos}(k(r - 1) + \phi)}{r^3} \\
U_\textrm{LJG} &= \frac{1}{12} - \frac{2}{r^6} - \epsilon \, \text{exp}\left(-\frac{(r - r_0)^2}{2\sigma^2}\right) .
\end{split}
\end{align}

Systems of 4096 particles are slowly cooled from a high temperature $ 1.0 \epsilon_0/k_B$ to a lower temperature $0.1 \epsilon_0/k_B$ over $10^8$ timesteps of size $\delta t = 0.005 t_0$ each, where $\epsilon_0$ and $t_0$ are the fundamental energy and time scales used for simulation, and $k_B$ is the Boltzmann constant. Pairwise potential parameters $k$, $\phi$, $\epsilon$, and $r_0$ for the self-assembly trajectories used in this work are listed in Table~\ref{tab:potential_params} and full trajectories are available in the Supplementary Information.

\begin{table}[h]
\caption{Pairwise interaction potential parameters for each of the self-assembled simulation trajectories used in this work.}
\label{tab:potential_params}
\centering \begin{tabular}{lcccc}\toprule
& \multicolumn{2}{c}{OPP} & \multicolumn{2}{c}{LJG}
\\\cmidrule(lr){2-3}\cmidrule(lr){4-5}
Structure & $k$ & $\phi$ & $r_0$ & $\epsilon$ \\ \midrule
$cF4$-Cu & $5$ & 2.8 & & \\
$tP30$-CrFe & $8.5$ & $1.5$ & & \\
$cP54$-K$_4$Si$_{23}$ & $8.5$ & $4$ & & \\
Icosahedral quasicrystal & $7.5$ & $3.9$ & & \\
$cI2$-W & & & $1.1$ & $3$ \\
$hP2$-Mg & & & $1.8$ & $0.1$ \\
\bottomrule
\end{tabular}
\end{table}

\section{Training hyperparameters}
\label{app:hyperparameters}

All networks used in this work produce and consume rotation-invariant values $v_i$ with 32 dimensions. MLPs $\mathcal{V}$, $\mathcal{S}$, and $\mathcal{R}$ have one hidden layer of 64 dimensions and ReLU activation. All optimization is carried out using the Adam~\cite{kingma_adam_2015} optimizer with a batch size of 4 point clouds and gradient accumulation over 16 batches. For static datasets (consisting of the autoencoding, frame identification, and nearest bond tasks), 30\% of the data are held back for the validation set. The learning rate is reduced by a factor of 0.5 at every epoch during which the validation loss does not decrease, and training is halted after 2 epochs of no validation loss improvement, or 128 total epochs. For dynamic datasets (consisting of the denoising, noisy bond identification, and shift identification tasks), 2048 batches are counted as a training epoch and 512 batches as a validation epoch. The learning rate is reduced by a factor of 0.5 after 4 epochs without improvement in the validation set loss, and training is terminated after 10 epochs without improvement, or 128 total epochs.

\section{Transfer learning performance}
\label{app:xfer_tables}

We list the key metrics from Figure~\ref{fig:transfer_learning} below for each data fraction (minor x axis) in Tables~\ref{tab:xfer_0}--\ref{tab:xfer_4}.

\begin{table}[h]
\caption{Transfer learning performance for models trained on a source task (indicated by column) for a target task (indicated by row) for a data ratio of 0.0001. For each task, final-layer retraining is shown above, while full-network fine-tuning is shown below. }
\label{tab:xfer_0}
\centering \tabcolsep=0.11cm \begin{tabular}{ccccccc}\toprule
 & Frame & Noisy & Autoencoder & Denoising & Nearest & Shift
\\ \midrule
\multirow{2}{*}{ Frame } & 0.98 & 0.984 & 0.97 & 0.957 & 0.962 & 0.972
\\
 & 0.99 & 0.968 & 0.97 & \textbf{ 0.956 } & 0.959 & 0.965
\\ \midrule
\multirow{2}{*}{ Noisy } & 0.312 & \textbf{ 0.306 } & 0.314 & 0.317 & 0.314 & 0.308
\\
 & \textbf{ 0.306 } & \textbf{ 0.306 } & \textbf{ 0.306 } & \textbf{ 0.306 } & \textbf{ 0.306 } & \textbf{ 0.306 }
\\ \midrule
\multirow{2}{*}{ Autoencoder } & 0.274 & 0.08 & \textbf{ 0.0115 } & 1.88 & 1.82e+06 & 0.936
\\
 & 0.0466 & 0.0293 & 0.0119 & 0.226 & 1.51e+05 & 0.0849
\\ \midrule
\multirow{2}{*}{ Denoising } & 0.000231 & 0.000254 & 0.000139 & 0.0018 & 0.0225 & 0.000195
\\
 & 0.000179 & 4.6e-05 & \textbf{ 5.09e-07 } & 1.12e-05 & 0.0225 & 8.39e-06
\\ \midrule
\multirow{2}{*}{ Nearest } & 0.14 & 0.141 & 0.185 & 0.144 & 0.0349 & 0.132
\\
 & 0.15 & 0.139 & 0.171 & 0.154 & \textbf{ 0.0339 } & 0.118
\\ \midrule
\multirow{2}{*}{ Shift } & 0.0282 & 0.0271 & 0.0335 & 0.0297 & 0.0253 & \textbf{ 0.0143 }
\\
 & 0.0266 & 0.0282 & 0.0347 & 0.0266 & 0.024 & 0.0189
\\ \bottomrule
\end{tabular} \end{table}
\begin{table}[h]
\caption{Transfer learning performance for models trained on a source task (indicated by column) for a target task (indicated by row) for a data ratio of 0.001. For each task, final-layer retraining is shown above, while full-network fine-tuning is shown below. }
\label{tab:xfer_1}
\centering \tabcolsep=0.11cm \begin{tabular}{ccccccc}\toprule
 & Frame & Noisy & Autoencoder & Denoising & Nearest & Shift
\\ \midrule
\multirow{2}{*}{ Frame } & 0.69 & 0.885 & 0.953 & 0.967 & 0.708 & 0.758
\\
 & 0.692 & 0.81 & 0.921 & 0.913 & \textbf{ 0.664 } & 0.728
\\ \midrule
\multirow{2}{*}{ Noisy } & 0.189 & \textbf{ 0.18 } & 0.208 & 0.185 & 0.191 & 0.184
\\
 & \textbf{ 0.18 } & \textbf{ 0.18 } & \textbf{ 0.18 } & \textbf{ 0.18 } & \textbf{ 0.18 } & \textbf{ 0.18 }
\\ \midrule
\multirow{2}{*}{ Autoencoder } & 0.268 & 0.0852 & 0.0126 & 1.67 & 1.82e+06 & 2.21
\\
 & 0.0449 & 0.0194 & \textbf{ 0.009 } & 0.0135 & 1.09e+03 & 0.0444
\\ \midrule
\multirow{2}{*}{ Denoising } & 0.000148 & 0.000194 & 0.000172 & 0.00481 & 432 & 0.000329
\\
 & 3.41e-05 & 3.09e-06 & \textbf{ 3.66e-07 } & 4.7e-06 & 0.0553 & 5.29e-06
\\ \midrule
\multirow{2}{*}{ Nearest } & 0.0883 & 0.0858 & 0.107 & 0.104 & \textbf{ 0.0215 } & 0.0809
\\
 & 0.0859 & 0.0853 & 0.103 & 0.0938 & 0.0236 & 0.0562
\\ \midrule
\multirow{2}{*}{ Shift } & 0.0255 & 0.0246 & 0.0267 & 0.0266 & 0.0216 & \textbf{ 0.0113 }
\\
 & 0.0176 & 0.0166 & 0.0196 & 0.0163 & 0.0144 & 0.0141
\\ \bottomrule
\end{tabular} \end{table}
\begin{table}[h]
\caption{Transfer learning performance for models trained on a source task (indicated by column) for a target task (indicated by row) for a data ratio of 0.01. For each task, final-layer retraining is shown above, while full-network fine-tuning is shown below. }
\label{tab:xfer_2}
\centering \tabcolsep=0.11cm \begin{tabular}{ccccccc}\toprule
 & Frame & Noisy & Autoencoder & Denoising & Nearest & Shift
\\ \midrule
\multirow{2}{*}{ Frame } & 0.494 & 0.721 & 0.763 & 0.716 & 0.507 & 0.577
\\
 & 0.51 & 0.572 & 0.496 & 0.459 & \textbf{ 0.39 } & 0.408
\\ \midrule
\multirow{2}{*}{ Noisy } & 0.169 & \textbf{ 0.159 } & 0.188 & 0.177 & 0.174 & 0.164
\\
 & \textbf{ 0.159 } & \textbf{ 0.159 } & \textbf{ 0.159 } & \textbf{ 0.159 } & \textbf{ 0.159 } & \textbf{ 0.159 }
\\ \midrule
\multirow{2}{*}{ Autoencoder } & 0.261 & 0.103 & 0.0144 & 1.02 & 1.8e+06 & 1.89
\\
 & 0.0309 & 0.0276 & 0.0125 & 0.01 & 29.4 & \textbf{ 0.000523 }
\\ \midrule
\multirow{2}{*}{ Denoising } & 9.93e-05 & 0.000389 & 0.000226 & 0.00115 & 0.339 & 0.00432
\\
 & 6.17e-05 & 2.98e-05 & \textbf{ 6.05e-07 } & 8.53e-06 & 58.1 & 1.35e-05
\\ \midrule
\multirow{2}{*}{ Nearest } & 0.0689 & 0.0661 & 0.077 & 0.0826 & 0.0174 & 0.0537
\\
 & 0.0602 & 0.048 & 0.0564 & 0.0504 & \textbf{ 0.0172 } & 0.0289
\\ \midrule
\multirow{2}{*}{ Shift } & 0.0248 & 0.0242 & 0.025 & 0.0264 & 0.0214 & \textbf{ 0.0111 }
\\
 & 0.0143 & 0.0127 & 0.0128 & 0.0127 & 0.0116 & 0.0115
\\ \bottomrule
\end{tabular} \end{table}
\begin{table}[h]
\caption{Transfer learning performance for models trained on a source task (indicated by column) for a target task (indicated by row) for a data ratio of 0.1. For each task, final-layer retraining is shown above, while full-network fine-tuning is shown below. }
\label{tab:xfer_3}
\centering \tabcolsep=0.11cm \begin{tabular}{ccccccc}\toprule
 & Frame & Noisy & Autoencoder & Denoising & Nearest & Shift
\\ \midrule
\multirow{2}{*}{ Frame } & 0.389 & 0.517 & 0.627 & 0.575 & 0.409 & 0.499
\\
 & 0.401 & 0.318 & 0.256 & 0.243 & 0.185 & \textbf{ 0.181 }
\\ \midrule
\multirow{2}{*}{ Noisy } & 0.168 & 0.158 & 0.183 & 0.175 & 0.175 & 0.162
\\
 & \textbf{ 0.157 } & \textbf{ 0.157 } & \textbf{ 0.157 } & \textbf{ 0.157 } & \textbf{ 0.157 } & \textbf{ 0.157 }
\\ \midrule
\multirow{2}{*}{ Autoencoder } & 0.25 & 0.109 & 0.0145 & 1.74 & 1.83e+06 & 2.23
\\
 & 0.00246 & 0.00827 & 0.0127 & \textbf{ 1.37e-05 } & 0.734 & 0.000654
\\ \midrule
\multirow{2}{*}{ Denoising } & 2.13e-05 & 3.26e-07 & 1.64e-06 & 0.00237 & 339 & 0.000738
\\
 & 0.0016 & 1.68e-05 & \textbf{ 4.46e-08 } & 2.46e-06 & 0.00643 & 2.45e-06
\\ \midrule
\multirow{2}{*}{ Nearest } & 0.0528 & 0.0533 & 0.06 & 0.0681 & 0.0155 & 0.0387
\\
 & 0.0423 & 0.0321 & 0.0353 & 0.0243 & \textbf{ 0.0153 } & 0.0191
\\ \midrule
\multirow{2}{*}{ Shift } & 0.0247 & 0.0242 & 0.0251 & 0.026 & 0.0215 & 0.0111
\\
 & 0.0144 & 0.0126 & 0.0123 & 0.0123 & 0.0112 & \textbf{ 0.0109 }
\\ \bottomrule
\end{tabular} \end{table}
\begin{table}[h]
\caption{Transfer learning performance for models trained on a source task (indicated by column) for a target task (indicated by row) for a data ratio of 1.0. For each task, final-layer retraining is shown above, while full-network fine-tuning is shown below. }
\label{tab:xfer_4}
\centering \tabcolsep=0.11cm \begin{tabular}{ccccccc}\toprule
 & Frame & Noisy & Autoencoder & Denoising & Nearest & Shift
\\ \midrule
\multirow{2}{*}{ Frame } & 0.346 & 0.388 & 0.508 & 0.429 & 0.408 & 0.48
\\
 & 0.322 & 0.238 & 0.142 & 0.157 & 0.106 & \textbf{ 0.0911 }
\\ \midrule
\multirow{2}{*}{ Noisy } & 0.194 & 0.196 & 0.184 & 0.175 & 0.164 & 0.162
\\
 & 0.187 & 0.215 & 0.163 & \textbf{ 0.158 } & 0.21 & 0.208
\\ \midrule
\multirow{2}{*}{ Autoencoder } & 0.265 & 0.115 & 0.0161 & 1.74 & 2.65e+04 & 2.11
\\
 & 0.0367 & 0.035 & 0.015 & 8.61e-06 & 0.000145 & \textbf{ 5.37e-07 }
\\ \midrule
\multirow{2}{*}{ Denoising } & 0.0021 & 0.0021 & 0.00211 & 0.00853 & 0.0462 & 0.0229
\\
 & \textbf{ 0.00207 } & 0.0168 & \textbf{ 0.00207 } & 0.00209 & 0.0041 & 0.00208
\\ \midrule
\multirow{2}{*}{ Nearest } & 0.0439 & 0.0459 & 0.0515 & 0.0551 & 0.0143 & 0.0311
\\
 & 0.0265 & 0.0221 & 0.0156 & 0.0151 & \textbf{ 0.0135 } & 0.0139
\\ \midrule
\multirow{2}{*}{ Shift } & 0.0248 & 0.0242 & 0.0251 & 0.0257 & 0.0215 & 0.011
\\
 & 0.0141 & 0.0127 & 0.0122 & 0.0124 & 0.0112 & \textbf{ 0.0109 }
\\ \bottomrule
\end{tabular} \end{table}

\section{Embedding snapshot study details}
\label{app:featurization_baselines}

\microsection{Baseline featurization methods.} As baselines for comparison, we generate featurizations based on the Steinhardt order parameters $Q_\ell$~\cite{steinhardt_bond-orientational_1983}, local neighbor-averaged spherical harmonics $|\Psi|$~\cite{spellings_machine_2018}, and a simple featurization based on radial distances. The Steinhardt order parameters are defined for a given spherical harmonic degree $\ell$ and set of near-neighbors; for our featurization, we calculate a vector of Steinhardt order parameters for a range of spherical harmonic degrees $\ell \in [2, 12]$. To account for the variety of neighborhoods that could be relevant, we also compute the order parameters using a range of nearest neighborhood sizes, from the nearest 4 to 20 particles. The local neighborhood spherical harmonics $|\Psi|$ create a rotation-invariant representation using the inertia tensor of each local environment and use the same neighbor accumulation process for each set of spherical harmonics, from 4 to 20 particles. The normalized radial distance featurization serves as a proxy for the use of radial distribution functions in materials science, which are probability distributions of finding pairs of particles at a given distance. In this case, we simply take the magnitude of each nearest-neighbor bond---up to 20 nearest neighbors, sorted by radial distance---and normalize by the nearest neighbor distance to make the representation scale-invariant, as the other featurizations are. This simple representation should be able to distinguish phase transitions and ordered structures, although it may have difficulties in some cases where the structures being compared are highly similar. The embedding methods used, as well as their dimensionality, are listed in Table~\ref{tab:embedding_dimensionality}.

\begin{table}[h]
\caption{Dimensionality of embedding methods used for Table~\ref{tab:zero_shot_aucs}.}
\label{tab:embedding_dimensionality}
\centering \begin{tabular}{lcc}\toprule \\
Method Abbreviation & Method Name & Dimensionality \\ \midrule
Frame & Frame Classification & 32 \\
Noisy & Noisy Bond Classification & 32\\
Autoencoder & Autoencoder & 8 \\
Denoising & Denoising Autoencoder & 32 \\
Nearest & Nearest Bond Regression & 32 \\
Shift & Shift Identification & 32 \\
$Q_\ell$ & Steinhardt $Q$ Vector & 102 \\
$|\Psi|$ & Local Neighborhood-averaged Spherical Harmonics & 2873 \\
$\frac{|r_n|}{|r_0|}$ & Radial Distance Vector & 19 \\
\bottomrule
\end{tabular}
\end{table}

\microsection{Snapshots used.} Details of systems used in the zero-shot phase distinction study described in Table~\ref{tab:zero_shot_aucs} are listed here.

\begin{enumerate}
\item $L_A/L_A$: Two liquids at the same temperature that will later form $cF4$-Cu. This system is a control experiment: identical liquids should be difficult or impossible to distinguish to a significant degree.
\item $G_A/L_A$: A gas and liquid that will later form $cI2$-W. This models a simple phase transition between low-density and high-density disordered phases.
\item $L_A/S_A$: A liquid that forms $cI2$-W and $cI2-W$ solid. This models a simple disorder-order transition.
\item $G_A/S_A$: A gas that will form an icosahedral quasicrystal and the quasicrystal itself. This models a disorder-order transition to a complex ordered structure.
\item $L_A/L_B$: Two liquids that will form $cF4$-Cu and $hP2$-Mg later. These solid structures are very similar, having identical representations when using nearest-neighbor graphs.
\item $S_A/S_B$: Two solid structures of $cF4$-Cu and $hP2$-Mg. Most methods developed to distinguish solid structures should be able to handle this baseline.
\end{enumerate}

\end{document}